\documentclass[11pt]{article}

\usepackage{subcaption,tikz}
\usetikzlibrary{spy}
\usepackage{fourier}
\usepackage{color}
\usepackage[usestackEOL]{stackengine}
\usepackage{graphicx}
\usepackage{url}
\usepackage[affil-it]{authblk}
\usepackage{graphicx} 
\usepackage{amsmath}
\usepackage{wrapfig}
\usepackage{comment}
\usepackage[T1]{fontenc}
\usepackage{times}
\usepackage{pgfplots}
\begin{document}

\title{Context Aware Object Geotagging}
\author{Chao-Jung Liu$^{\diamond}$, Matej Ulicny$^{\diamond}$, Michael Manzke$^{\diamond}$ \& Rozenn Dahyot$^{\star}$}
\affil{ADAPT Research Centre, $^{\diamond}$Trinity College Dublin, $^{\star}$Maynooth University, Ireland}

\date{}
\maketitle
\thispagestyle{empty}

\begin{abstract}
Localization of street objects from images has gained a lot of attention in the recent years. We propose an approach to improve asset geolocation from street view imagery by enhancing quality of the metadata associated with the images using Structure from Motion. The predicted object geolocation is further refined by imposing contextual geographic information extracted from OpenStreetMap. Our pipeline is validated experimentally against the state of the art approaches for geotagging traffic lights.
\end{abstract}

\section{Introduction}
\begin{wrapfigure}{r}{0.5\textwidth}
  \vspace{-20pt}
  \begin{center}
    \includegraphics[width=0.48\textwidth]{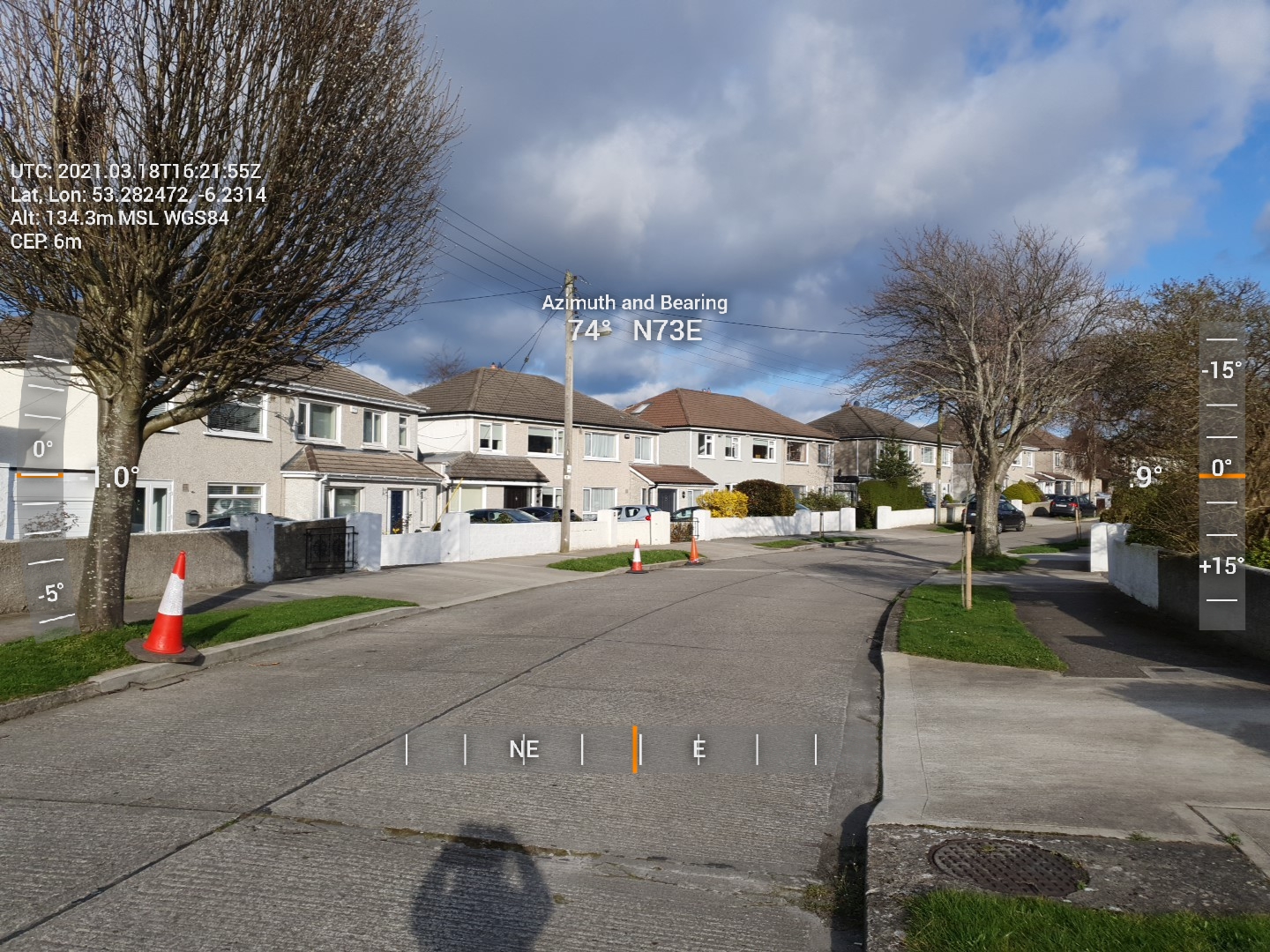}
\end{center}
\vspace{-20pt}
  \caption{Example of a street view image with its metadata overlaid. The image is captured with Dioptra app.} \label{fig:image:meta}
  \vspace{-10pt}
\end{wrapfigure}
Monitoring public assets is a labour-consuming task and for many decades, solutions collecting street view imagery have been routinely deployed in combination with computer vision-based approaches for object detection and recognition in images \cite{Dahyot01}. 

Nowadays, street view images are available in massive amounts (e.g.: Mapillary\footnote{\label{footnote:mapillary}https://www.mapillary.com/}, Google Street View (GSV)\footnote{https://developers.google.com/maps/documentation/streetview/overview}) and additional information about the scene can be further extracted by machine learning techniques.
Krylov et al.~\cite{KrylovICIP18,KrylovRS18} have employed deep learning modules for segmenting objects of interest (e.g. poles) in images and estimating their distance from the camera, and a Markov Random Field (MRF) is then used as a decision module to provide a usable list of the GPS coordinates of the assets of interest, limiting duplicates by reconciling detection from multiple view images.
  
\begin{figure}[ht!]
	\centering
	\includegraphics[width=\textwidth]{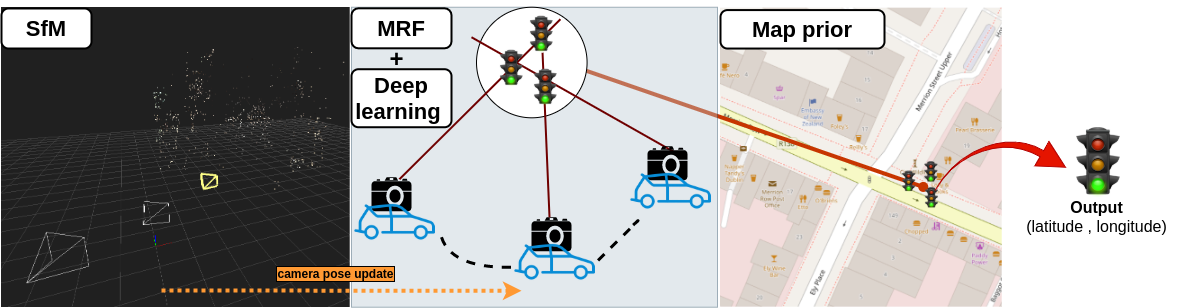}
	\caption{Pre-processing: SfM aims to de-noise camera medadata (i.e. poses) used as an input of the MRF. Post-processing: the Map prior module  refines the result from MRF using contextual information from OSM.}
	\label{fig:pipeline_fusion}
\end{figure}

The MRF conveniently merges information extracted from images and their metadata i.e. their associated camera location (GPS) and bearing information (cf. Fig.~\ref{fig:image:meta}).
\stepcounter{footnote}
\footnotetext{\label{footnote:dioptra}https://play.google.com/store/apps/details?id=com.glidelinesystems.dioptra{\scriptsize\footnotetext{https://play.google.com/store/apps/details?id=com.glidelinesystems.dioptra}}}\hspace{-1.4em} Currently, the pipeline of Krylov et al. assumes that the metadata associated with the camera view pose is noiseless, however, it is not always the case (e.g. due to GPS receiver imprecision) and consequently, this noise affects the accuracy of the geo-location of the assets found. In this paper, we propose to improve that pipeline by (1) denoising the camera metadata using Structure from Motion (SfM) and (2) using contextual information extracted from Open Street Map (OSM)\footnote{https://www.openstreetmap.org/} to push the predictions to a more probable area where the objects should be situated based on road and building locations. Fig.~\ref{fig:pipeline_fusion} summarizes our contributions and our approach has been validated for traffic light geolocation (c.f. Sec.\ref{sec:Results}).

\section{State of the Art}
\subsection{Camera geolocalization}
Various Simultaneous Localization and Mapping (SLAM) and SfM techniques have been proposed to infer 3D points and to estimate the motion from a set of images \cite{klingner2013street,torii2009google,cummins2009highly,lee2009robust,heinly2015reconstructing}. Bundle adjustment (BA) is integrating matched points within a sequence of images and finding a solution simultaneously optimal with respect to both camera parameters and 3D points. Agarwal et al.~\cite{agarwal2010bundle} is first to propose the bundle adjustment that is used in the structure from motion. The trajectory of camera pose estimation is based on relative measurements, error accumulation over time thus leads to drift.
Lhuillier~\cite{lhuillier2011fusion} proposed to use GPS geotag in the bundle adjustment optimization. 
A similar problem is the camera re-localization~\cite{yu2016monocular,agarwal2015metric}. A GPS tag and SfM technique are used to geo-localize a street view image by estimating its relative pose against images from a database.
Bresson et al.~\cite{bresson2019urban} and Kendall et and al.~\cite{kendall2015posenet} proposed to employ a CNN (Convolution Neural Network) features to estimate camera pose transformation.

\subsection{Object geotagging}

Qin et al.~\cite{qin2019tlnet,qin2019monogrnet} proposed to estimate the instance-level depth of objects in images as an alternative to pixel-wise depth estimation. They found out the latter (obtained by minimising the mean error for all pixels) sacrifices the accuracy of certain local areas in images. 
Bertoni et al.~\cite{bertoni2019monoloco} employed a prior knowledge of the average height of humans to perform pedestrian localization.
Qu et al.~\cite{qu2015vehicle} proposed to detect and locate traffic signs from a monocular video stream. They relied on bundle adjustment with image GPS geo-tag to reconstruct a sparse point cloud as a 3D map, then align it with several landmarks from the 3D city model generated by Soheilian et al. \cite{soheilian2013generation}.

Wegner et al.~\cite{wegner2016cataloging} proposed a probabilistic model to locate trees. They employed multiple modalities, including aerial view semantic segmentation, street view detection, map information as well as the tree distance prior. Information is fused into a conditional random field (CRF) to predict the positions of trees. However, identical features may be mismatched in case the recurring objects sit nearby. To solve this issue, Nassar et al.~\cite{nassar2019learning,nassar2020multi} employed the soft geometry constraint on geo-location of camera pose to identify a same object that appears in two views.  They concatenate camera pose information together with image features and decode them using a CNN. The same object in first view can be re-identified in the second view.     

Nassar et al.~\cite{nassar2020geograph} extend the method by constructing a graph from detected bounding boxes across the multi-views, feed the graph to a GNN~\cite{kipf2016semi} and let the GNN identify the same objects across different views.
Hebbalaguppe. et al.~\cite{hebbalaguppe2017telecom} predicted bounding boxes around street objects, which was followed by the two-view epipolar constraint to reconstruct 3D feature points from the two observed scenes. However, the 3D feature point does not necessarily fall inside the target bounding box. Krylov et al.~\cite{KrylovRS18} employed the camera pose from multiple views as a soft constraint and used semantic segmentation of images alongside a monocular depth estimator to extract the information (bearing and depth) about objects of interest, and feeds the obtained information into an MRF model that predicts their locations.



\section{Methods}
\label{sec:our:work}

We present camera calibration using the SfM technique in Section~\ref{sfm}, which provides a higher quality information to be used as an input to the MRF presented in Section~\ref{data_prep}. Section~\ref{MRF_postpro} proposes a post-processing method to refine the MRF predictions.

\begin{figure}
    \centering
    \includegraphics[width=\textwidth]{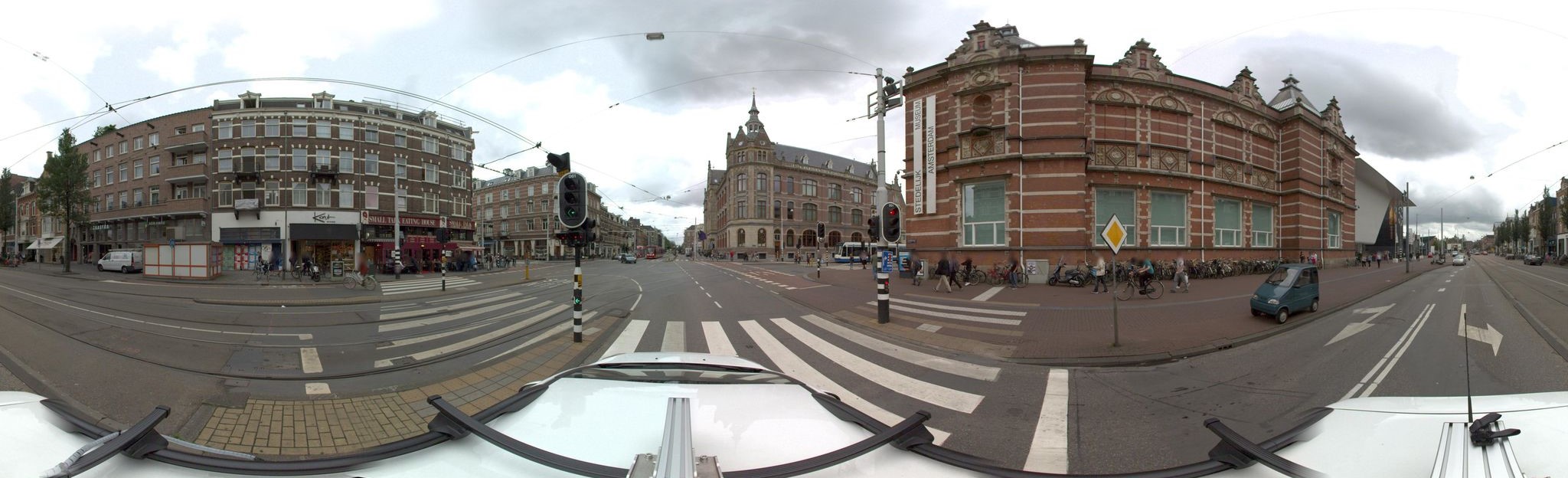}
    \hfill
    \centering
    \includegraphics[width=0.12\textwidth]{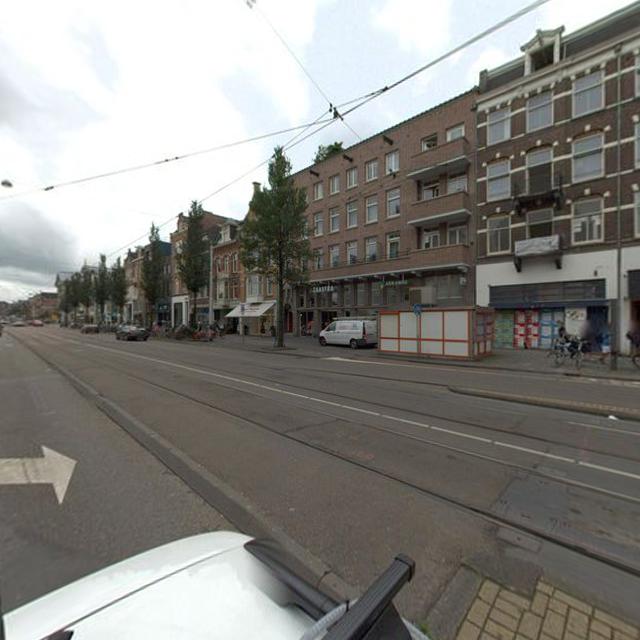}
    \includegraphics[width=0.12\textwidth]{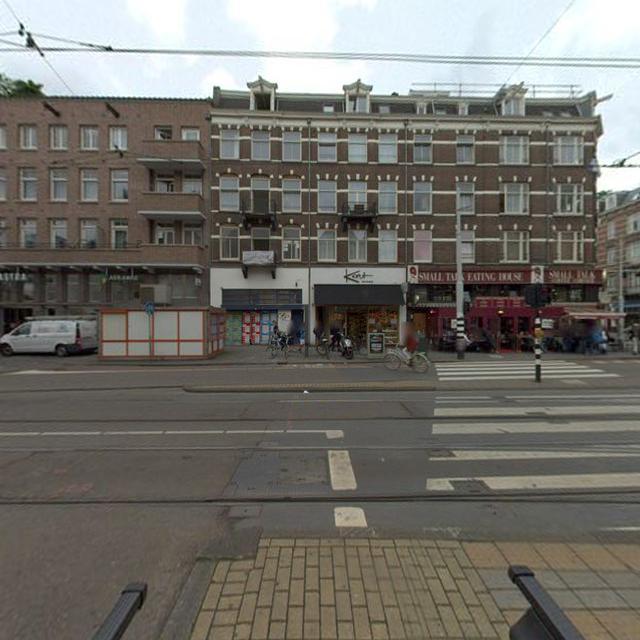}
    \includegraphics[width=0.12\textwidth]{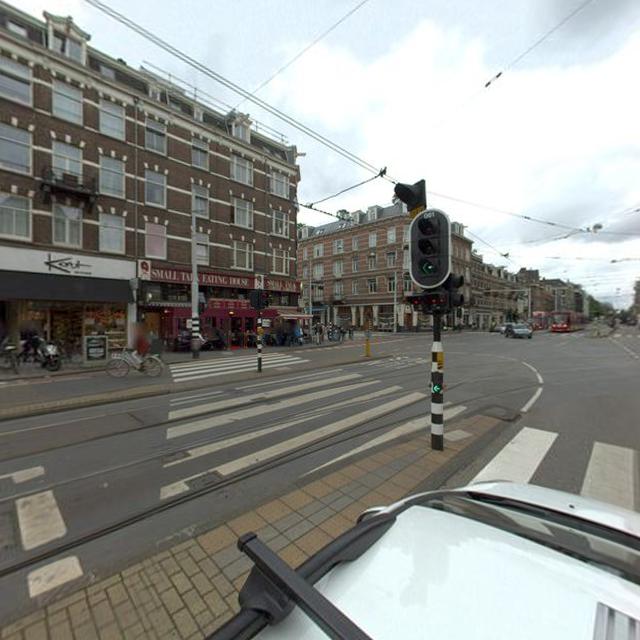}
    \includegraphics[width=0.12\textwidth]{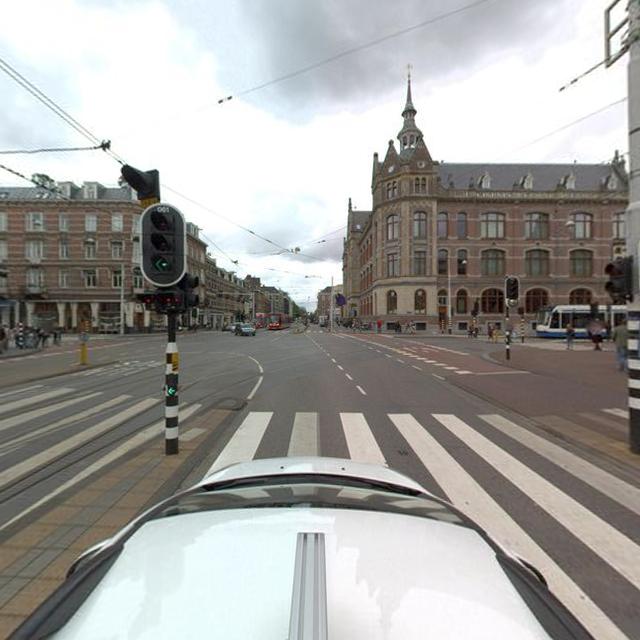}
    \includegraphics[width=0.12\textwidth]{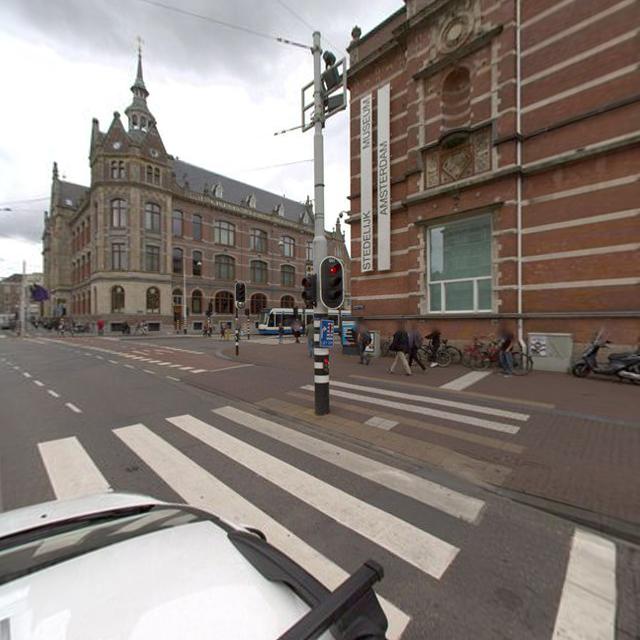}
    \includegraphics[width=0.12\textwidth]{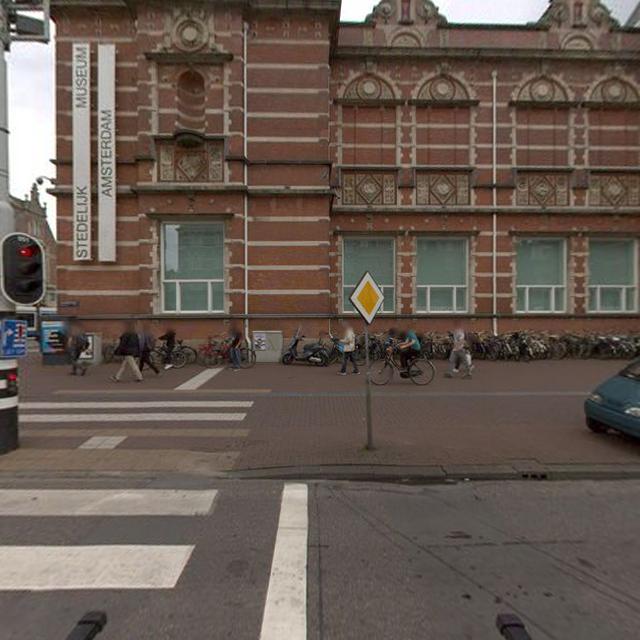}
    \includegraphics[width=0.12\textwidth]{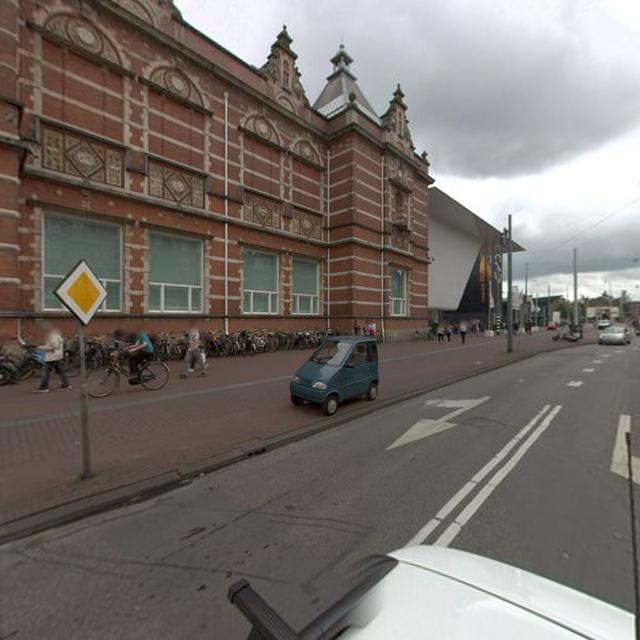}
    \includegraphics[width=0.12\textwidth]{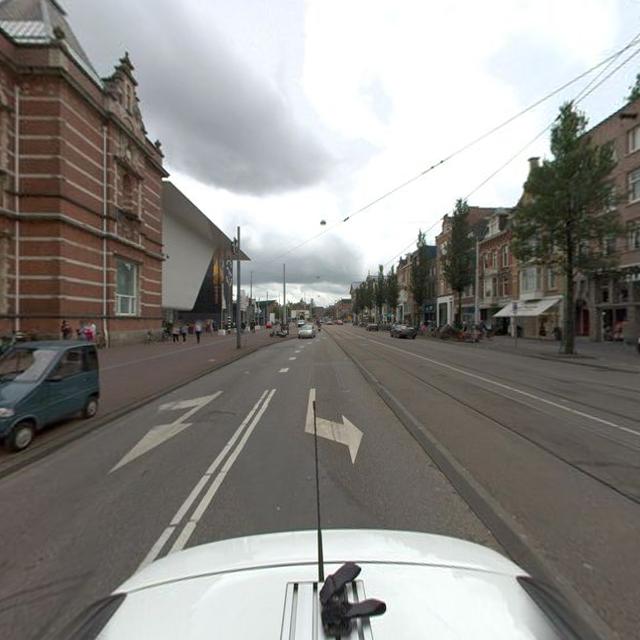}
    \caption{Input image representation consists of 8 overlapping rectilinear views split from a 360° field of view panorama. The image shown above is acquired from Mapillary API.}
    \label{fig:input_images}
\end{figure}

\subsection{Structure from Motion: using optical observation to denoise on GPS data}
\label{sfm}

The input represents a set of $N$ panoramic street view images (360° field of view) captured with their metadata in an area of interest. 
Accurate camera geo-location is a key to accurately geo-locate objects in the scene. The GPS position in the metadata is inherently noisy, which lowers the accuracy for predicting object positions. 
To get a better estimate of the GPS coordinates associated with each camera position, we propose to tune each of the camera positions with a conventional 3D reconstruction pipeline~\cite{andrew2001multiple}, followed by bundle adjustment~\cite{agarwal2010bundle}.
To ease image matching, we split the 360° panorama views into 8 overlapping rectilinear views: each view covers a 90-degree field of view and is overlapped by 45 degrees in the horizontal direction. Each view is then considered as an image captured by a pinhole camera, free of distortion (see Figure~\ref{fig:input_images}). 

We aim to find all possible matching features extracted from our images and perform camera calibration to adjust the camera pose from image metadata. 

We note the set of rectilinear views 
$\mathcal{V} =  \lbrace v_1^{(i)}, \cdots, v_8^{(i)}\rbrace_{i=1,\cdots,N} $ where 
$ v_1^{(i)}, \cdots, v_8^{(i)}$ corresponds to  
rectilinear views associated with panorama $i=1,\cdots,N$ ($N=112$ in our experiment).
Suppose two views are matched by their detected features. The epipolar constraint with 5 point algorithm \cite{andrew2001multiple} is applied to find the essential matrix $\mathrm{E}$ which establishes the geometry relationship between two views. $\mathrm{E}$ can be further decomposed into translation and rotation matrix, noted as $\mathrm{R}$ and $\tau$, respectively. They can be put together as a transformation matrix $\Theta \in SE(3)$ 
\begin{equation}
\Theta =
\begin{pmatrix}
\mathrm{R} & \tau \\
0 & 1\\
\end{pmatrix} \in \mathbb{R}^{4\times 4} \quad \text{with}\quad \mathrm{R} \in SO(3) \ \text{and} \ \tau \in \mathbb{R}^{3}.
\end{equation}
Each calibrated view in $\mathcal{V}$ is associated with  $\Theta=(\mathrm{R},\tau)$, these parameters can be estimated by minimizing the re-projection error from 3D feature space to 2D image plane within a bundle of images.

\subsection{Object geolocation with MRF}
\label{data_prep}

The MRF model performs binary decisions on the nodes of a 2D graph, each node corresponding to an intersection between two rays.
 The rays correspond to rays (in 2D)  with origins the camera GPS coordinates and with directions the bearings associated with the segmented object of interest (the pixel in the middle of the segmented object is chosen for the bearing information). The objects of interest are segmented using a deep learning pipeline that also estimates their distances (from the camera) \cite{KrylovRS18}. 
Each camera view provides one or many rays shooting to the objects of interest.
The MRF model is optimised to perform a binary decision for each node concerning its occupancy by the object of interest (i.e. 0 = no object, 1 = object present). 
For more information, please refer to \cite{KrylovRS18}. Our contribution in this paper is in providing more accurate GPS coordinates for the camera positions (than originally available in the image metadata) thanks to SfM,  hence improving the geo-location of the nodes on this MRF and ultimately improving the accuracy of GPS coordinates for the objects of interest.

\subsection{Post-processing}
\label{MRF_postpro}

\begin{figure}[h!]
	\centering
	\includegraphics[width=0.49\linewidth]{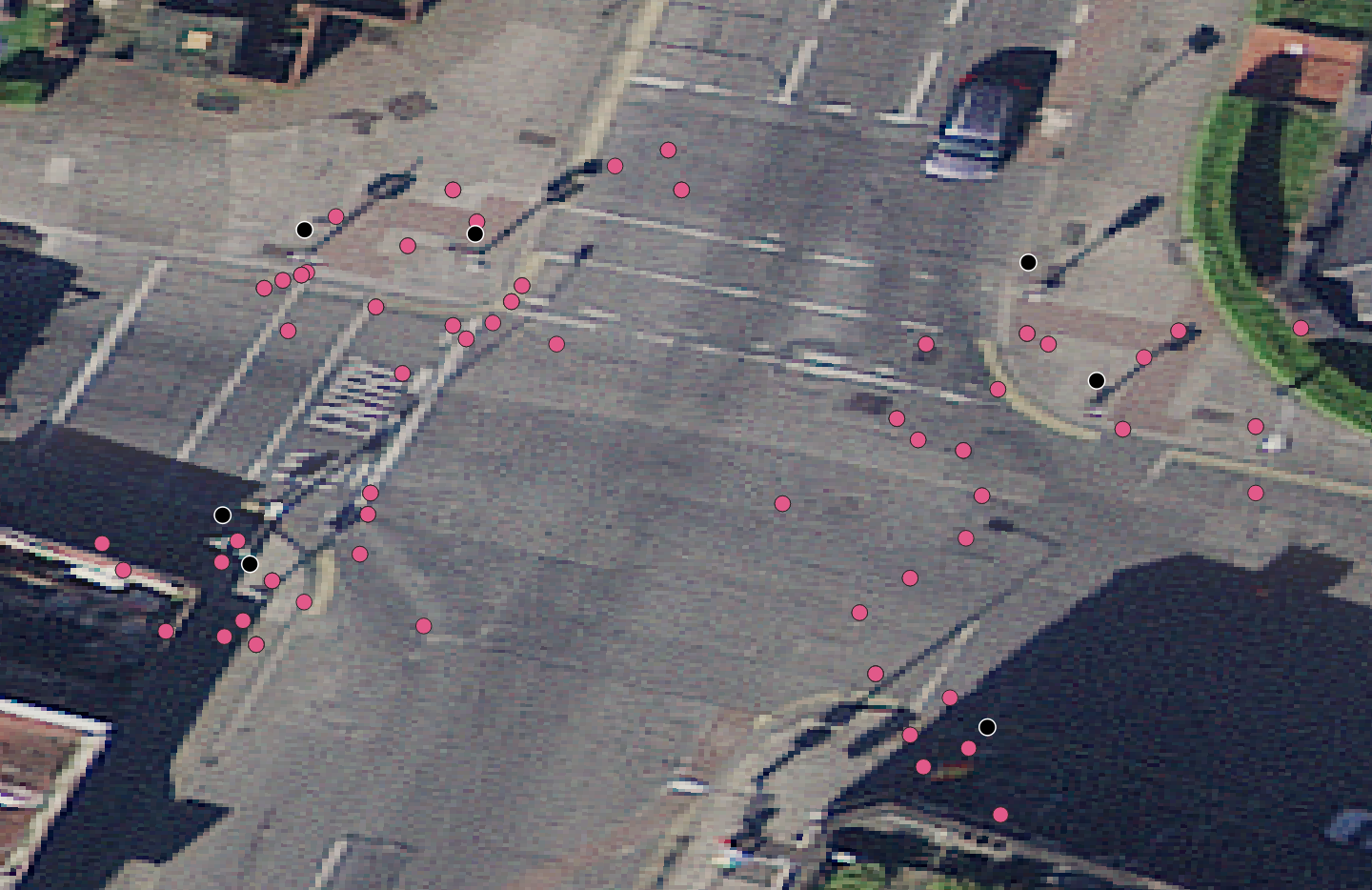}
	\includegraphics[width=0.49\linewidth]{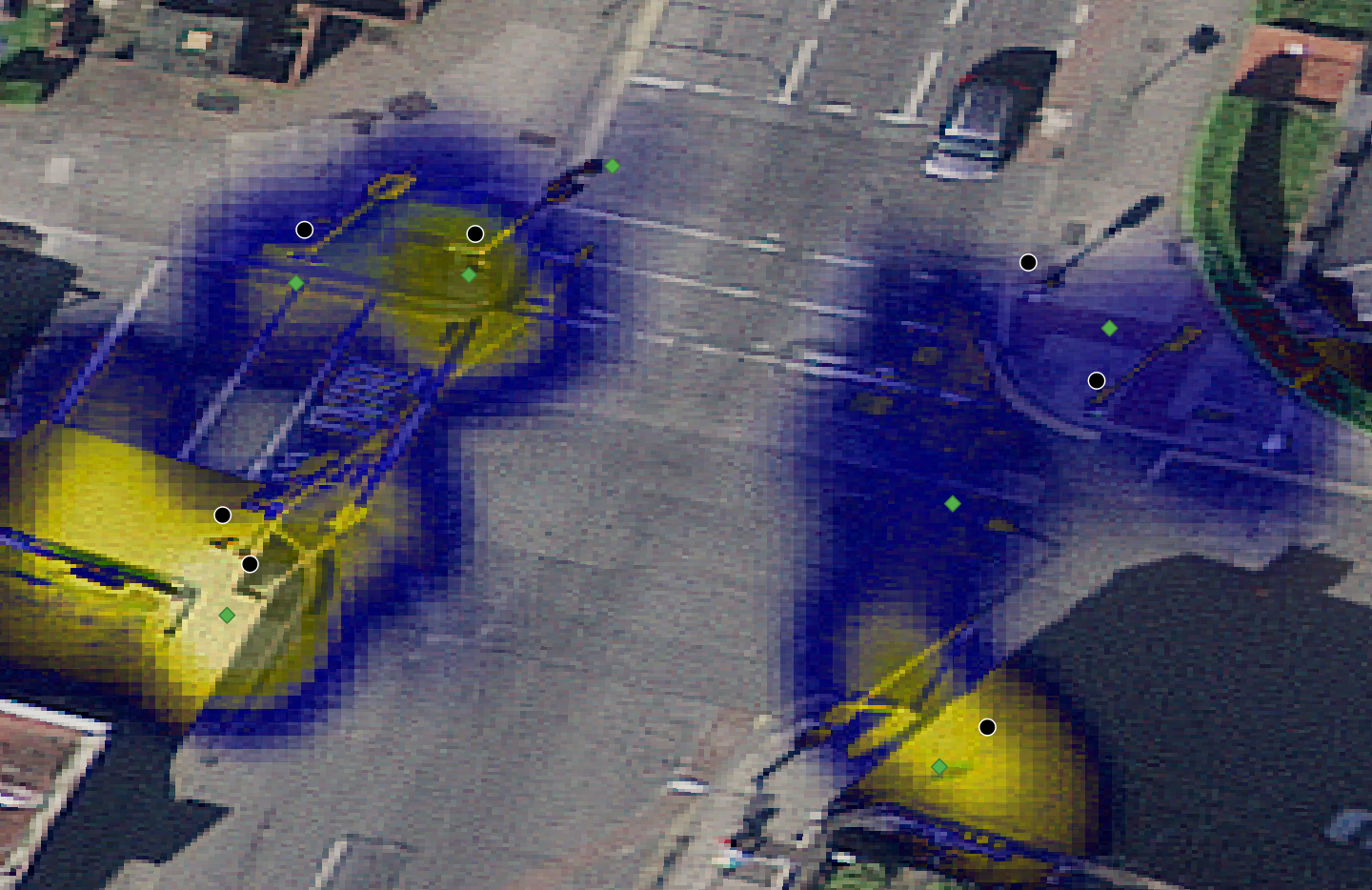}
	\caption{On the left figure, the red dots are positive intersections from MRF. The location of ground truth are shown in black dots. On the right figure, the green dots are the result after clustering process. Probability density function is applied to demonstrate the points' density of the intersections. The color code from blue to yellow means the number of intersections from small to many.}
	\label{fig:MRF_output}
\end{figure}

Because of the inaccuracies of the rays that define the MRF nodes, the same object may be associated with multiple nodes (Fig.~\ref{fig:MRF_output} left) located in the same vicinity on the MRF graph. To resolve this issue Krylov et al.~\cite{KrylovRS18} added a hierarchical clustering step after optimising the MRF to merge close positive sites together. The final position is the average of sites in the cluster. 
However, we have observed that some of the positive sites were situated at improbable areas, for example, in the middle of the road. Therefore, we propose here to use OSM data to act as a useful prior for an area. As our objects of interest (e.g. traffic lights) are static and are located on a side of the road, we apply the following rule: \textit{the object of interest can not be located in the middle of the road, or around the edge of a building}. The OSM data has building and road classes represented by polygons and lines, respectively. A Normal kernel is fitted at each OSM node (cf. Fig.~\ref{Fig:aerial}). Suppose a cluster $C$ containing $n$ positive sites $C=[c_{1},c_{2},...,c_{n}]$, $W(x)$ is the function to query the weight that corresponds to the particular site in $C$ and depends on the  OSM nodes $N_x$ within the close proximity of the site $x$. The position $P$ (equation~\ref{Eq:weighted_average}) can be refined using a weighted average where certain sites are penalized with small weights. 
\begin{equation}
W(x) = 1 - \min\left(1,\sum_{\mu,\sigma\in N_x} - \exp\left(\frac{ - \left(x - \mu  \right)^2} {2\sigma^2} \right) \right)
\quad \text{and} \quad
P = \frac{\sum_{i=1}^{n} W(c_{i}) \times c_{i}}{\sum_{i=1}^{n} W(c_{i})}
\label{Eq:weighted_average}
\end{equation} 
The $\sigma$ stands for the standard deviation in meters and the $\mu$ is the node centre obtained from the OSM data. The $\sigma$ is set to 2 meters for roads and 1 meter for building edges. The resolution of the Gaussian grid is 25 centimetres.

\subsection{Implementation}

SURF descriptors \cite{bay2008speeded} are used and in each view, 6,000 features are extracted. We employ FLANN (Fast Library for Approximate Nearest neighbours \cite{muja2012fast})  to match the SURF features between rectilinear views. RANSAC is used to remove outliers. We use the OpenSfM~\footnote{https://github.com/mapillary/OpenSfM} to calibrate camera poses and Crese~\cite{agarwal2010bundle} as our solver to optimize the $\Theta$. As the nodes from raw OSM data are not equally distributed, we interpolate the nodes every 5 meters in QGIS~\footnote{https://www.qgis.org/en/site/} to get a dense distribution of the map prior. 




\section{Experimental Results}
\label{sec:Results}

\begin{figure}[!h]
\begin{tabular}{cc}
\resizebox{0.5\textwidth}{!}{
\begin{tikzpicture}[spy using outlines={circle,yellow,magnification=3.5,size=9.5cm, connect spies}]
\node (a) at (0,0){\includegraphics[interpolate=true,width=1.2\textwidth,height=17.2cm]{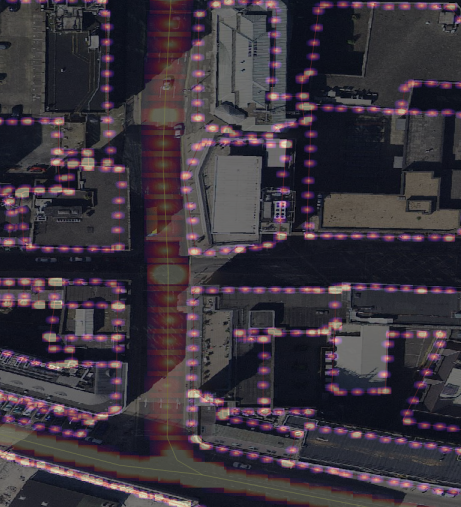}};
\spy on (-2.3,-1.6) in node [left] at (8.2, 3.0);

\node (b) at (a.south)[anchor=north,xshift=-0.85cm]{
\begin{axis}[
    hide axis,
    scale only axis,
    height=0pt,
    width=0pt,
    colormap={vioyellow}{color=(yellow) color=(red)
    color=(violet)},
    colorbar horizontal,
    point meta min=0,
    point meta max=1,
    colorbar style={
        width=\textwidth,
        xtick={0,0.1,0.2,...,1},
        height = 0.5cm
    }]
    \addplot [draw=none] coordinates {(0,0)(0,0)};
    \end{axis}
    };
\end{tikzpicture}
}
&
\resizebox{0.48\textwidth}{!}{
\includegraphics[width=\textwidth]{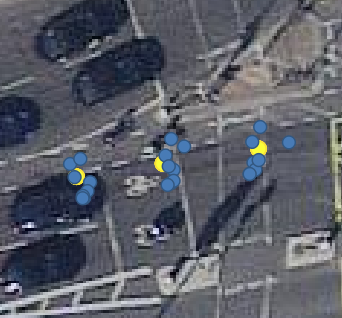}
}
\\
\end{tabular}
\caption{The figure (\textbf{left}) shows the prior map information overlaid on an aerial image. The normal kernel is applied to each node that is imported from OSM. The heatmap outlines improbable object locations that will have a smaller contribution towards the weighted sum.
On the (\textbf{right}) yellow dots are the positions taken from image metadata and the blue dots represent their corrected versions with SfM.}
\label{Fig:aerial}
\end{figure}

To validate our approach, we have used 896 GSV images (112 panoramas split into 8 images each) 
collected in Dublin city centre. The object of interest corresponds to a traffic light. We fine-tuned the input camera poses using the SfM (cf. Fig.~\ref{Fig:aerial}). Our method corrected the bearing and position information on average by 4.36 degrees and 0.71 meters respectively. 
Moreover, the use of the OSM prior results in an average refinement of the prediction by 0.17 meters.
\begin{table}[!h]
	\begin{center}
		\begin{tabular}{cccccccc}
		$\#$Actual&$\#$Detected&TP&Precision$\uparrow$ &Recall$\uparrow$ & F-measure$\uparrow$&\Longstack{Geo-localization \\ error$\downarrow$}&\Longstack{Geo-localization \\ error$\downarrow$(with OSM)}\\
		
			\hline
			\hline
			\multicolumn{8}{c}{no correction \cite{KrylovRS18}}\\
			\hline
		76&94&58&0.61&0.76&0.68&2.71&2.64\\
		
		\hline
		\multicolumn{8}{c}{correction on $\tau$ only}\\
		\hline
		76&89&57&0.64&0.75&0.69&2.79&2.74\\
		
		\hline
		\multicolumn{8}{c}{correction on $\mathrm{R}$ and $\tau$}\\
		\hline
		76&92&54&0.57&0.72&0.64&2.53&2.48\\
		\hline
		\hline
		\end{tabular}
		
\end{center}
\vspace{-1.5em}
\caption{We evaluate the impact of metadata correction by a comparison with results that do not use any pose correction. By correcting the full camera pose ($\mathrm{R}$ and $\tau$), the geo-location accuracy reaches error of around 2.5 meters to a reference point. It outperforms the result with no correction by 18cm, and 16 cm after applying the OSM prior. We reach the highest F-measure if only the $\tau$ is corrected.}
\label{tab:result_1}
\end{table}
 
By using our SFM module we can check the impact of the following correction of the metadata: correction on $\tau$ only (i.e. GPS location of the camera), correction on $\mathrm{R}$ and $\tau$ (i.e. correction of both GPS location and bearing of the camera). To validate our approach, we use the original metadata as our baseline for comparisons.
Table~\ref{tab:result_1} shows the testing results in terms of geo-localization error and precision and recall detection metrics. We consider traffic lights to be recovered accurately (true positive) if they are located within 6 meters from the reference position, otherwise it is viewed as a false positive. 
The geo-localization error measures the average Haversine distance between the prediction and its reference target in meters. A small distance indicates accurate position prediction. 

We compare our results with related public asset geo-location approaches in Table~\ref{Table:comparing_approaches}. The proposed technique reaches the smallest positional error, however, the results are not directly comparable due to the different complexity of the scene and detected objects.

\hfill

\begin{table}[!h]
	\begin{center}
		\begin{tabular}{llcc}
			\hline
		\multicolumn{4}{c}{Comparison with other methods}\\
		\hline
		Method&Dataset&F-measure$\uparrow$&Geo-localization error$\downarrow$\\
		\hline
		Siamese CNN~\cite{nassar2020multi}&Pasadena~\cite{wegner2016cataloging}&0.51&3.13\\
		Siamese CNN&Mapillary~\cite{neuhold2017mapillary}&0.72&4.36\\
		GNN-Geo~\cite{nassar2020geograph}&Pasadena&0.64&2.75\\
		GNN-Geo&Mapillary&0.87&4.21\\
		\textbf{Ours}&\textbf{DTL}~\cite{KrylovRS18}&\textbf{0.64}&\textbf{2.48}\\
			\hline
		\end{tabular}
	\end{center}
	\vspace{-1.5em}
	\caption{In comparison with other approaches, our method achieves the smallest geo-localization error, although the other datasets might be more challenging for object detection.}
	\label{Table:comparing_approaches}
\end{table}

\section{Conclusion}

We have shown that by denoising metadata associated with street view imagery using SfM, and by using context information such as road and building shapes extracted from OSM,  assets of interest can be geolocalized with higher accuracy. Currently, our pipeline is geotagging one class of objects at a given time, and future work will investigate multiple static object class tagging with additional priors associated with their relative positioning in the scene, to improve further geolocation accuracy.

\section*{Acknowledgments}
This research was supported by the ADAPT Centre for Digital Content Technology funded under the SFI Research Centres Programme (Grant 13/RC/2106) and co-funded under the European Regional Development Fund. We gratefully acknowledge the support of NVIDIA Corporation with the donation of the Titan Xp GPUs.

\bibliographystyle{plain}
\bibliography{imvip}

\end{document}